\documentclass[letterpaper, 10 pt, conference]{ieeeconf}  

\IEEEoverridecommandlockouts                              

\usepackage{import}
\usepackage{color, colortbl}
\usepackage{amsmath}
\usepackage{mathtools,amssymb,commath}
\usepackage{graphicx,import}
\usepackage{url}
\usepackage{multirow}

\newcommand{\mx}[1]{\boldsymbol{#1}}

\newcommand{\vect}[1]{\boldsymbol{#1}}

\title{\LARGE \bf
Nonlinear In-situ Calibration of Strain-Gauge Force/Torque Sensors \\ for Humanoid Robots
}

\author{Hosameldin Awadalla Omer Mohamed$^{1, \hspace{0.5 mm} 2}$,Gabriele Nava$^{1}$,\\ Punith Reddy Vanteddu$^{1, \hspace{0.5 mm} 3}$, Francesco Braghin$^{2}$ and Daniele Pucci$^{1, \hspace{0.5 mm} 3}$ 
\thanks{This work was supported by Istituto Italiano di Tecnologia, 16163 Genova, Italy. (Corresponding author: H. A. O. Mohamed)}
\thanks{$^{1}$H. A. O. Mohamed, G. Nava, P. R. Vanteddu and D. Pucci are with the Artificial and Mechanical Intelligence, Istituto Italiano di Tecnologia, 16163 Genova, Italy {\tt\small firstname.surname@iit.it}}
    \thanks{$^{2}$H. A. O. Mohamed and F. Braghin are with the Department of Mechanical Engineering, Politecnico di Milano, 20156 Milan, Italy {\tt\small hosameldinawadalla.mohamed@polimi.it}}%
    \thanks{$^{3}$D. Pucci is also with the School of Computer Science, Univ. of Manchester, Manchester M13 9PL, U.K.}
}

\begin{document}

\maketitle
\thispagestyle{empty}
\pagestyle{empty}

\begin{abstract}

High force/torque (F/T) sensor calibration accuracy is crucial to achieving successful force estimation/control tasks with humanoid robots. State-of-the-art affine calibration models do not always approximate correctly the physical phenomenon of the sensor/transducer, resulting in inaccurate F/T measurements for specific applications such as thrust estimation of a jet-powered humanoid robot. This paper proposes and validates nonlinear polynomial models for F/T calibration, increasing the number of model coefficients to minimize the estimation residuals. The analysis of several models, based on the data collected from experiments with the iCub3 robot, shows a significant improvement in minimizing the force/torque estimation error when using higher-degree polynomials. In particular, when using a 4th-degree polynomial model, the Root Mean Square error (RMSE) decreased to 2.28N from the 4.58N obtained with an affine model, and the absolute error in the forces remained under 6N while it was reaching up to 16N with the affine model.

\end{abstract}


\section{INTRODUCTION}
\label{sec:introduction}

Humanoid robots have the capability to perform various tasks such as locomotion, manipulation, and interaction with the surroundings~\cite{goswami2019humanoid}, and even possibly to perform aerial locomotion~\cite{nava2018position}. Usually, humanoid robots are endowed with different sensors to help them achieve the aforementioned tasks. Such as the estimation of external forces (contact forces, thrust forces). Relatively large-scale humanoid robots have multiple six-dimension force/torque sensors, either mounted inside the kinematic chains of their limbs~\cite{Bartolozzi2017icub}, or on their extremities~\cite{stasse2017talos, negrello2016walkman}.\looseness=-1

The most common force/torque sensing technology is based on strain-gauges that exploit the piezo-resistive effect~\cite{barlian2009piezoresistance}. 
These sensors consist of multiple strain-gauges that produce small electrical signals that represent the applied forces/torques, which are then amplified and sampled to be read by a computer system. The calibration procedure consists in finding the relationship between the readings, often referred to as the \emph{raw signals}, and the exerted forces/torques. It is performed by applying known forces/torques, possibly measured with another calibrated F/T sensor, and recording the raw signals that change as a result of these forces/torques.\looseness=-1
When discussing the calibration models of the F/T sensors, usually the raw signals are referred to as the \emph{input}, and the forces/torques as the \emph{output} of the system, although in terms of causality the flow direction is reversed.\looseness=-1

\begin{figure}[t]
	\centering
	\includegraphics[width=1\linewidth]{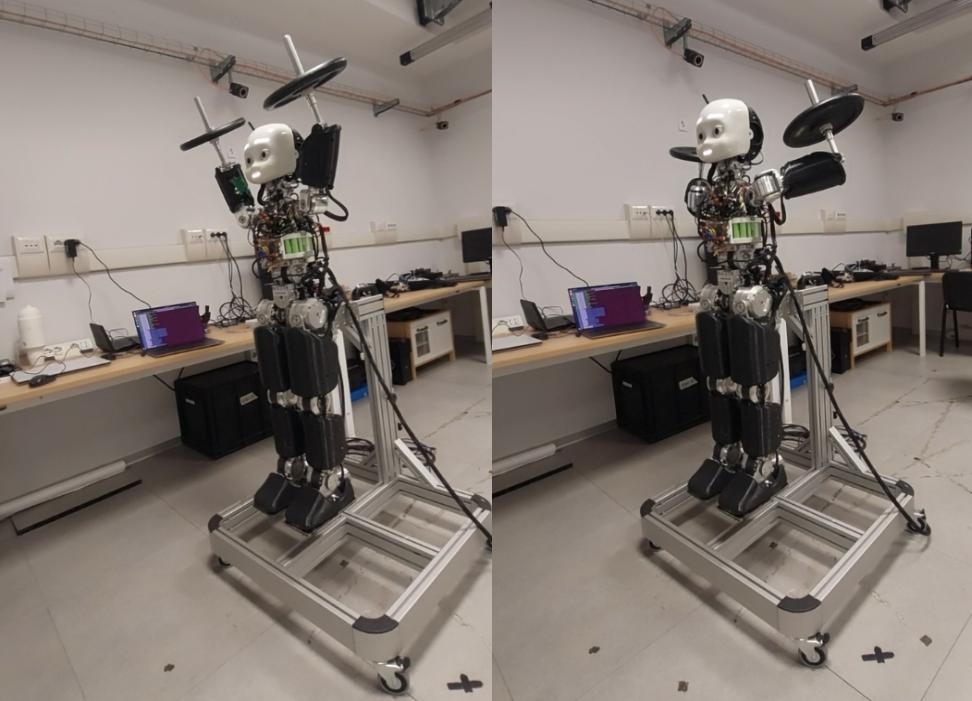}
	\caption{Pictures showing the robot's configurations taken while executing the \emph{Weight Lifting trajectory}, defined in Section~\ref{sec:experiments}.}
	\label{fig:lifting_configs}
\end{figure}

It has been investigated and verified that calibrating the F/T sensors using a dedicated setup and then mounting them on the robot, referred to as \emph{off-situ} calibration, can invalidate the identified model and hence decrease the quality of the measurement~\cite{chavez2019insitu}.\looseness=-1

Moreover, strain-gauge F/T sensors have a considerable dependency on the temperature, because it can change their resistance and experience drifts in the measured forces/torques~\cite{barlian2009piezoresistance}. For this reason, some manufacturers define specific operating ranges 
in which the measurements are valid~\cite{ati20xxdatasheet}; others add an internal temperature sensor to approximate the temperature values of each strain gauge, and include the temperature measurements in the calibration model~\cite{robotiqs2019datasheet}. Further, the authors in~\cite{chavez2019temperature} performed \emph{in-situ} calibration taking also the internal temperature into account.\looseness=-1

F/T sensors mounted on the robot's limbs have many different applications. The particular application that motivated this work is estimating the thrust intensities for a flying jet-powered humanoid robot. In a previous work, a thrust estimation approach was presented, although the upper-body F/T sensors were not directly exploited to estimate the turbines thrust~\cite{Mohamed2022thrust}.\looseness=-1

The main drawback was that the measurement accuracy when using state-of-art affine calibration models ~\cite{chavez2019insitu,traversaro2015insitu,ding2022insitu} was not sufficient for the aforementioned application, since the ranges of applied forces/torques are much higher than during calibration. Moreover, the operating conditions are more severe concerning temperature variations, as the flight experiments are usually performed outdoors. In addition, the sensors are close to the jet engines and the joint motors, which generate heat because of the high load they sustain. Accurate modeling of the temperature effect is therefore essential for achieving acceptable performances.\looseness=-1
%
%

The advantage of affine calibration models is that they minimize the cross-coupling effects between different sensor outputs, but they do not catch
the deformations generated by the applied force/torque which results in nonlinear outputs~\cite{oh2018deeplearning, su2019modelfree}, therefore a part of the sensor's characteristics cannot be fully represented with the affine model.\looseness=-1

Many approaches in the literature already used model-free tools to obtain better calibration models with Neural Networks~\cite{su2019modelfree, oh2018deeplearning, tienfu1997neural}. However, Neural Networks need careful design of the training dataset, and require large data points to avoid the risk of overfitting. These requirements are challenging for in-situ calibration in the operating conditions of our experiment.\looseness=-1

Therefore, this paper proposes the design of different calibration models, 
based on polynomials, through which we increase the model capacity to represent possible nonlinear features and temperature effects while still preserving the interpretability of our models, and the possibility to understand
which are the higher-degree terms that capture the nonlinearities of the sensor.\looseness=-1

The remainder of the paper is organized as follows. Section~\ref{sec:modeling} describes the models that will be used to calibrate the F/T sensor, Section~\ref{sec:calibration_algorithm} explains the formulation of the optimization problem for training the models. Section~\ref{sec:experiments} details the experimental setup used to perform the in-situ calibration, and section~\ref{sec:results} validates and compares the performances of the proposed models. Finally, Section~\ref{sec:conclusions} presents final remarks and future directions.\looseness=-1

\section{FORCE/TORQUE SENSORS MODELING}
\label{sec:modeling}

Most of the calibration methods, irrespective of either \emph{in-situ} or \emph{off-situ}, utilize an affine model to describe the effect of strain-gauge F/T sensors. This model is motivated by laws of physics, namely, Hooke's law and piezo-resistive effect~\cite{barlian2009piezoresistance}. The model can be written as follows:\looseness=-1

\begin{equation}
\label{eq:lin_model}
    \vect{y} = \mx{C} \vect{u} + \vect{o},
\end{equation}

Where $\vect{y} \in \mathbb{R}^6$ is the applied wrench (force/torque) in [N] for the force components and in [Nm] for the torque components, $\vect{u} \in \mathbb{R}^6$ are the raw signals of the sensors (in bit counts), $\mx{C} \in \mathbb{R}^{6 \times 6}$ is referred to as the calibration matrix in [N(Nm)/bit] and $\vect{o} \in \mathbb{R}^6$ represents the offset.\looseness=-1

Instead, this paper proposes several nonlinear models to calibrate the F/T sensors. They are essentially polynomial models. More specifically:\looseness=-1

\begin{itemize}
    \item When considering polynomial degrees $>2$, we take into account also mixed products between different inputs, not only raising the inputs to a higher power.\looseness=-1 
    \item We allow including more inputs to the model, in addition to the row F/T sensor's measures. In this paper, we are considering adding temperature measure to the input $\vect{u}$.
\end{itemize}


Our F/T model is the combination of six different polynomial models, one for each force and torque component $y_i \in \mathbb{R}$, $i = (1,...,6)$. Each model uses the full input $\vect{u} \in \mathbb{R}^7$.\looseness=-1

To simplify testing of different models, we write an expression that generalizes all the model classes we would like to test. To achieve that, we define a few parameters that describe each model class. They are:

\begin{itemize}
    \item $np$: the degree of the polynomial.
    \item $ny$: number of outputs. In this work, it is set to 6, which is the dimension of expected F/T forces and torques $\vect{y}$.
    \item $nu$: number of inputs. In this work, it is set to 7, which is the dimension of actual F/T measurements, in addition to the internal temperature measurement.
\end{itemize}

\subsection{Higher-degree models ($np>1$)}

Re-writing the affine model \eqref{eq:lin_model} for each element of $y \in \mathbb{R}^6$ gives the following:

\begin{equation}
    \label{eq:np1}
    y_i(k) = \sum_{l=1}^{nu} \alpha_{il} \, u_l(k) + o_i ,
\end{equation}

Where $\alpha_{il}$ is the vector of coefficients that corresponds to the output component $y_i$, and $o_i$ is the coefficient that represents the corresponding offset.

The polynomial of degree $np=2$ includes the same terms of \eqref{eq:np1} plus a quadratic term, similarly for the polynomial of degree $np=3$, we include a cubic term, and so on. The polynomial model can be written as:

\begin{multline}
    \label{eq:pol_commutative}
    y_i(k) = \sum_{l=1}^{nu} \alpha_{il} \, u_l(k) + \underbrace{\sum_{l=1}^{nu} \sum_{m=1}^{nu} \beta_{ilm} \, u_l(k) u_m(k)}_{\text{Quadratic terms}} + \\ \underbrace{\sum_{l=1}^{nu} \sum_{m=1}^{nu} \sum_{q=1}^{nu} \gamma_{ilmq} \, u_l(k) u_m(k) u_q(k)}_{\text{Cubic terms}} + ... + o_i ,
\end{multline}

For polynomials of degree $np>1$, we don't need to consider all the possible terms: we can discard the commutative terms since they will not be identifiable, for example, the term $\beta_{i23} u_2(k) u_3(k)$ will not be distinguishable from the term $\beta_{i32} u_3(k) u_2(k)$ as the identification algorithm will not be able to uniquely identify $\beta_{i23}$ and $\beta_{i32}$. Moreover, discarding the commutative terms will also reduce the number of elements of the regressor matrix used in the identification procedure.

Therefore, by replacing the commutative terms with only the combinations, we obtain the following model:

\begin{multline}
    \label{eq:pol_combinations}
    y_i(k) = \sum_{l=1}^{nu} \alpha_{il} \, u_l(k) + \underbrace{\sum_{l,m \in \mathcal{G}} \beta_{ilm} \, u_l(k) u_m(k)}_{\text{Quadratic terms}} + \\ \underbrace{\sum_{l,m,q \in \mathcal{K}} \gamma_{ilmq} \, u_l(k) u_m(k) u_q(k)}_{\text{Cubic terms}} + ... + o_i ,
\end{multline}
where $i= (1, \dots, ny)$. $\mathcal{G}$ is the set of all the possible \emph{combinations} of $l$ and $m$ \emph{with repetition}. The number of quadratic terms equals
\begin{equation}
    \label{eq:no_combinations}
    \frac{(nu+r-1)!}{(nu-1)(r!)},
\end{equation}

In this case $r=2$. Also, $\mathcal{K}$ is the set of all the possible combinations of $l$, $m$ and $q$ (with repetitions). The number of terms in this can can be also obtained from \eqref{eq:no_combinations} by setting $r=3$. And so on for the rest of the terms.

For example, concerning the quadratic terms, instead of considering all the possible terms, for $nu=6$:
\begin{multline}
    l,m = ((1,1),(1,2),...,(1,6),(2,1),(2,2),...,(2,6),...,\\(6,1),(6,2),...,(6,6)),
\end{multline}

which are in total $6 \times 6 = 36$ terms. Considering only the combinations of $l$ and $m$ with repetitions, leads to the following terms
\begin{multline}
    l,m = ((1,1),(1,2),...,(1,6),\\(2,2),(2,3),...,(2,6),(3,3),(3,4),...,(3,6),\\(4,4),(4,5),(4,6),(5,5),(5,6),(6,6)),
\end{multline}

which are in total $\frac{(6+2-1)!}{(6-1)(2!)} = 21$ terms.

Applying the same example for the cubic terms case ($r=3$) will result in $6 \times 6 \times 6 = 216$ terms when adopting the model \eqref{eq:pol_commutative}, and the number of terms reduces to $\frac{(6+3-1)!}{(6-1)(3!)} = 56$ terms when including only the combinations (using \eqref{eq:no_combinations}).

\subsection{Model complexity}

We represent the \emph{model complexity} by the number of model coefficients to be identified, which are the values of $\alpha$, $\beta$, $\gamma$ and the offset $o$. The model complexity varies by changing the values of $np$, $ny$, $nu$. 
For example, a 3rd-degree polynomial model has values $np=3$, $ny=6$, and $nu=7$. This model has $720$ coefficients to be estimated. The model complexity rises rapidly when increasing the polynomial degree to $np=4$, the number of coefficients increases to $1980$. In testing F/T nonlinear models, we shall define a trade-off between model complexity and the accuracy of the calibration.\looseness=-1



\section{CALIBRATION ALGORITHM}
\label{sec:calibration_algorithm}

The model presented in \eqref{eq:no_combinations} can be rewritten as a linear system of equations of the form:

\begin{equation}
    \mx{A} \vect{x} = \vect{b} ,
\end{equation}
whose dimensions depend on the values of $np$, $ny$ and $nu$. The matrix $\mx{A} = \mx{A}(\vect{u})$ is the regressor matrix, the vector $\vect{x} = \vect{x}(\alpha, \beta, \gamma, o)$ contains the coefficients to be identified, and $\vect{b} = \vect{b}(\vect{y})$ Contains the output terms. Note that this is done for each output component $y_i$.


The value of the $\vect{x}$ is determined by solving the following optimization problem:
\begin{equation}
    \label{eq:qp}
    \vect{x}^* = \underset{\vect{x}}{argmin} \frac{1}{2} ||\mx{A} \vect{x}-\vect{b}||^2  .
\end{equation}

OSQP solver is used to solve the optimization problem~\cite{stellato2020osqp}.

\subsection{Reducing the model complexity with $l^1$-norm regularization}
\label{subsec:algorithm_lasso}

An $l^1$-norm regularization term weighted with the parameter $\lambda$ is added to the objective function to enforce sparsity of the obtained solution~\cite{tibshirani1996lasso}. This is particularly important for high $np$, because increasing the model complexity adds many terms that might have a negligible contribution. The regularization helps, in this case, to force their corresponding coefficients to be close to $0$. The $l^1$-norm regularization, known as LASSO~\cite{tibshirani1996lasso}, can be written as:

\begin{equation}
    \label{eq:qp_lasso}
    \vect{x}^* = \underset{\vect{x}}{argmin} \frac{1}{2} ||\mx{A} \vect{x}-\vect{b}||^2  + \lambda ||\vect{x}||_1  .
\end{equation}

To improve the performances and stability of the model, \emph{feature normalization} is also applied to have all features (columns of the regression matrix $\mx{A}$) on a similar scale.

\section{EXPERIMENTS}
\label{sec:experiments}

\begin{figure}[t]
	\centering
	\includegraphics[width=0.5\linewidth]{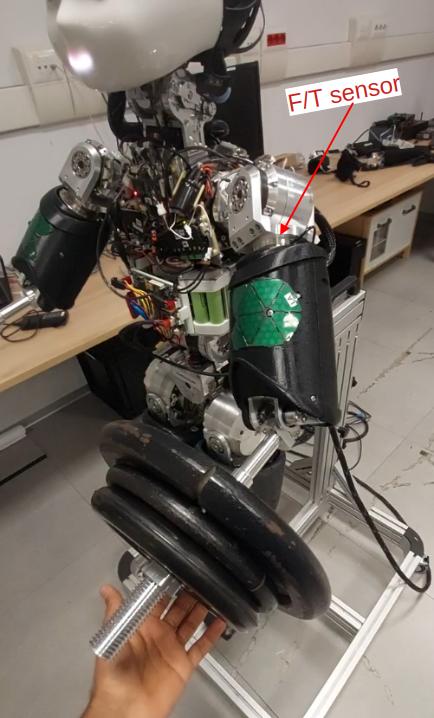}
    \caption{Location of the F/T sensor mounted on the arm of iCub3, highlighted with a red arrow. The image also shows the metal support used to attach the disk weights to the robot's forearms.}
	\label{fig:icub_ft_arm}
\end{figure}

The experimental setup consists of the humanoid robot iCub3~\cite{dafarra2022icub3}, which can also be endowed with jet engines and become iRonCub~\cite{pucci2017fly}. The robot contains eight internal six-axis F/T sensors. Four of them are mounted at the base of each limb, while four of them are mounted on the soles (front and rear for each sole) of the robot. Fig.~\ref{fig:icub_ft_arm} shows the location of the F/T sensor mounted on the arm of iCub3. The used F/T sensors are custom-made and developed by the iCub-Tech facility~\cite{icub-tech}. They contain 12 semiconductor strain gauges arranged in a six half-(Wheatstone)-bridges configuration. The F/T sensors studied in this work are of the model \text{FT45\textunderscore M4\textunderscore E2}.\looseness=-1

In this paper, we focus on the sensors mounted on the robot arms to perform in-situ experiments.
For modeling the robot, the Unified Robot Description Format (URDF) representation is used to describe the robot's kinematic and dynamic parameters~\cite{urdf}. 
In order to increase the range and variety of the forces/torques measured by the F/T sensor, we designed a support on which we can mount different weights. We used weight plates of \textbf{1kg}, \textbf{2kg}, and \textbf{5kg} to incrementally increase the load and reach \textbf{10kg} on the forearm. Fig~\ref{fig:icub_ft_arm} shows the metal support attaching \textbf{9kg} to the robot's forearm. Furthermore, we vary the temperature to reproduce the typical outdoor conditions of iRonCub experiments. The ranges of temperature, forces, and torques applied during the experiments are reported in Table \ref{tab:force_torque_ranges}.\looseness=-1

%

\begin{table}[t]
    \caption{Range of forces, torques and temperatures applied to the F/T sensor mounted on the robot's arm.}
\label{tab:force_torque_ranges}
\begin{center}
\begin{tabular}{ |c|c|c| } 
\hline
Signal & min value & max value \\
\hline
$F_x [N]$ & -30.45 & 123.91 \\
$F_y [N]$ & -120.48 & 34.03 \\
$F_z [N]$ & -113.24 & 113.65 \\
$T_x [Nm]$ & -4.35 & 26.99 \\
$T_y [Nm]$ & -4.68 & 25.55 \\
$T_z [Nm]$ & -12.59 & 7.27 \\
$Temp. [^{\circ}C]$ & 25 & 44 \\
\hline
\end{tabular}
\end{center}
\end{table}

\subsection{Calibration Trajectories}
\label{sec:trajectories}

In the in-situ calibration procedure, we exploit the gravity force, the known weights, and the robot model to calculate the expected force and torque values on the F/T. We modify the sensor's orientation by moving the arms, to enable the load to apply various wrenches that could cover the $\mathbb{R}^6$ space as much as possible. 
%
%
One issue is that the subspace of achieved forces and torques is limited by several factors. For instance, the set of joint configurations the robot can perform is limited by the joint limits and the necessity to avoid self-collisions. In addition, the maximum weight the robot can hold is reduced for some arm configurations.\looseness=-1


The trajectories used to train and validate the calibration models consist of the following types:

\subsubsection{Grid trajectory}

\begin{figure}[t]
	\centering
	\includegraphics[width=1\linewidth]{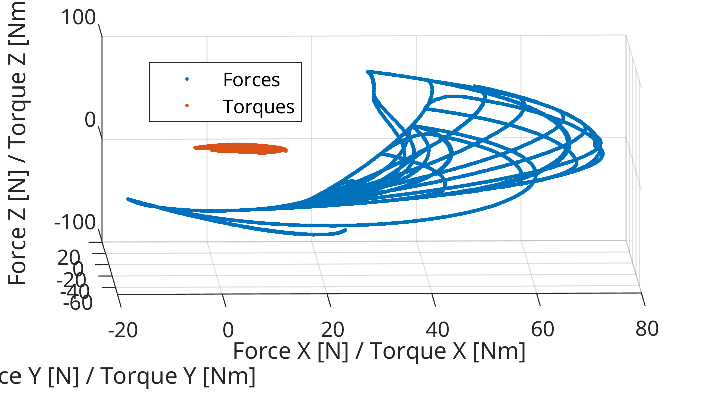}
	\caption{Plot of the 3D force components together with the 3D torques components for the \emph{Grid} trajectory. The forearms of the robot are removed and replaced with a load of approximately 2kg weight.}
	\label{fig:grid_traj}
\end{figure}

\begin{figure}[t]
	\centering
	\includegraphics[width=1\linewidth]{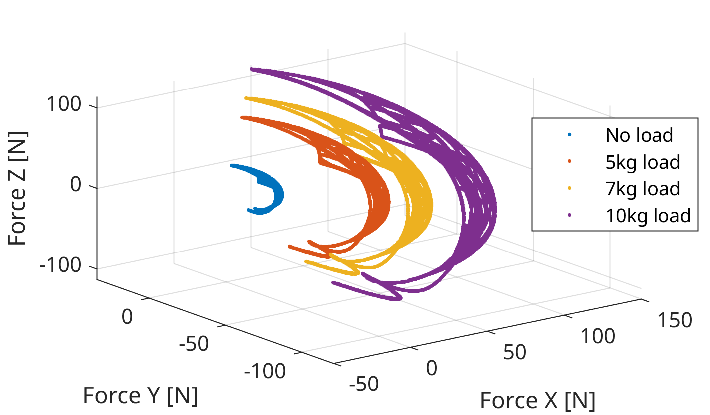}
	\caption{Plot of the 3D force components for the \emph{Grid} trajectory performed with loads of different weights. Each color corresponds to the weight attached to the robots's forearms.}
	\label{fig:grid_traj_multiple_loads}
\end{figure}

In this trajectory, the 3-dimensional plotting of the forces shows a grid-like shape that looks like a quarter of a sphere~\cite{guedelha2016self}. Fig. \ref{fig:grid_traj} shows the forces and torques measured on the F/T sensor while applying this trajectory.\looseness=-1
In order to cover a wider range, the same trajectory was performed with loads of different weights, attached to the forearm, ranging from \textbf{0kg} (only the metal support) to \textbf{10kg}. This enlarged the range of forces and torques measured with the Grid trajectory 
as shown by Fig.~\ref{fig:grid_traj_multiple_loads}. 
Attempting this type of trajectory with a load of more than \textbf{10kg} considerably increases the risk of damaging the robot, and then more safety precautions must be taken.\looseness=-1

\subsubsection{Weight Lifting trajectory} 
\begin{figure}[t]
	\centering
	\includegraphics[width=1\linewidth]{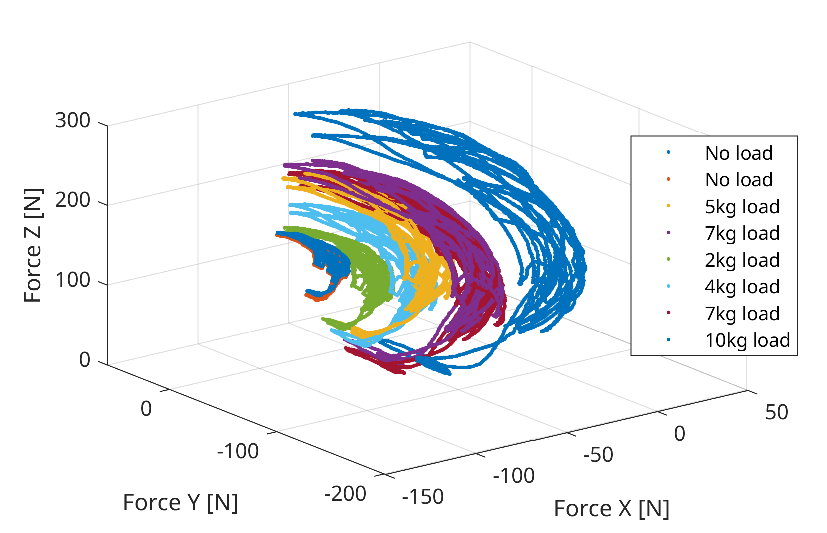}
    \caption{Plot of the 3D force components of the full trajectory used during the experiments. The full trajectory consists of two categories explained in section~\ref{sec:trajectories}: the \emph{Grid Trajectory} and the \emph{Weight Lifting Trajectory}. Each color corresponds to the weight attached to the robots's forearms.}
	\label{fig:full_traj_forces}
\end{figure}

\begin{figure}[t]
	\centering
	\includegraphics[width=1\linewidth]{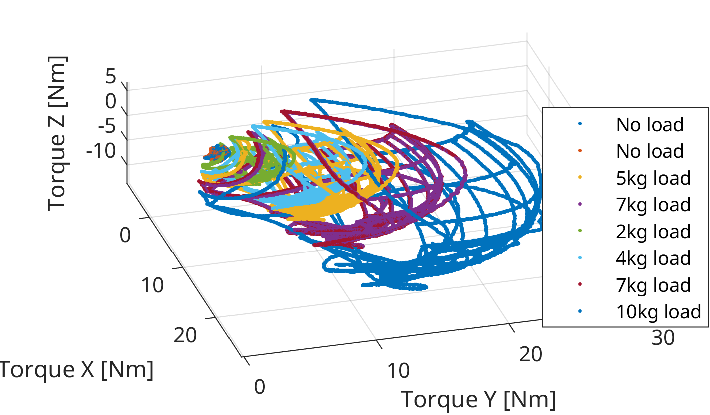}
	\caption{Plot of the 3D torque components of the full trajectory used during the experiments. The full trajectory consists of two categories explained in section~\ref{sec:trajectories}: the \emph{Grid Trajectory} and the \emph{Weight Lifting Trajectory}. Each color corresponds to the weight attached to the robots's forearms.}
	\label{fig:full_traj_torques}
\end{figure}

The robot lifts the load as high as possible, allowing the weight to excite the force component along the Z direction in the sensor's frame; then it moves the arm remaining close to the lifted configuration, and finally, it brings the arm down to release part of the load on the F/T sensor. 
%
%
This motion complements the Grid trajectory because it covers additional regions in the force/torque space. Moreover, it can be of particular importance in approximating the effect of jet engine's thrusts during flight tests. Shots of the trajectory are depicted in Fig. \ref{fig:lifting_configs}, while Figs. \ref{fig:full_traj_forces} and \ref{fig:full_traj_torques} show the full range of forces and torques collected with the two trajectories.\looseness=-1

\begin{figure}[t]
	\centering
	\includegraphics[width=1\linewidth]{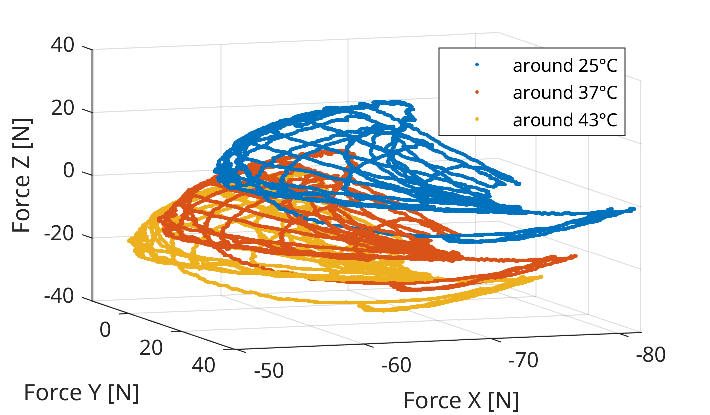}
	\caption{Plots of the 3D force components of the Grid trajectory. They are performed while carrying the same load, under different ambient temperatures. The color corresponds to the recorded internal temperature of the F/T sensor when performing the motion.}
	\label{fig:temperature_effect}
\end{figure}

We performed experiments at different temperature values. Based on the internal temperature sensor's reading, the covered range was from 25°C to 44°C. The temperature variation was achieved by performing the experiments in different conditions ranging from an air-conditioned room to under direct sun. Keeping the robot active for a long period also increases the F/T sensor's internal temperature due to the accumulation of the heat generated by the nearby motors and electronic boards. Fig. \ref{fig:temperature_effect} illustrates the effect of the temperature by comparing F/T measurements carrying the same load under different temperature values.\looseness=-1

\begin{figure}[t]
	\centering
        \includegraphics[width=0.7\linewidth]{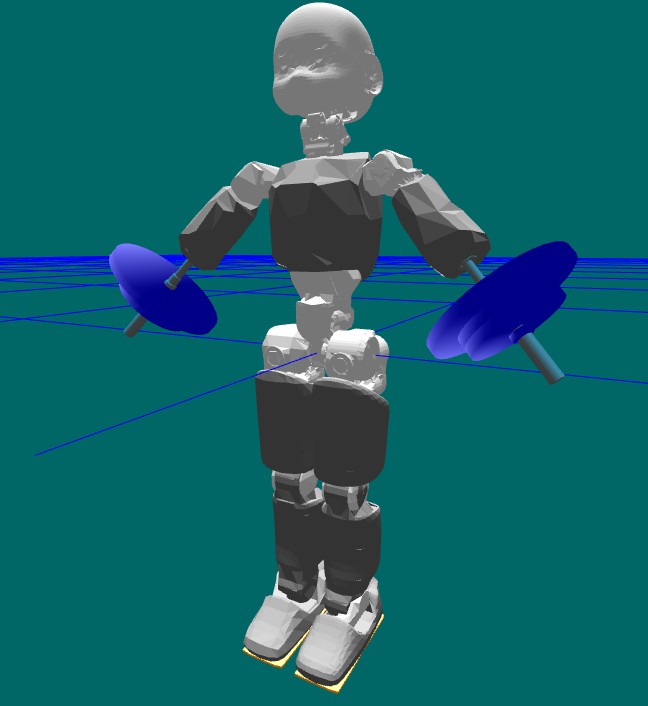}
	\caption{3D visualization of the robot model using urdf representation.}
	\label{fig:visualize_urdf}
\end{figure}

After collecting the data, we used iDynTree~\cite{nori2015idyntree} to parse the URDF models and compute the expected wrenches of each F/T. Fig. \ref{fig:visualize_urdf} shows the 3D visualization of the model URDF representation. The algorithm to compute the expected wrenches, is detailed in~\cite{traversaro2017modelling} (section 4.7), assuming perfect knowledge of the robot model and accurate joint position information. The measurements of the F/T sensors against the expected values represent the datasets used for training and validating the models detailed in~\ref{sec:modeling}, using the algorithm described in~\ref{sec:calibration_algorithm}.\looseness=-1

\section{CALIBRATION RESULTS}
\label{sec:results}

\subsection{Comparing polynomial models of various degrees}
\label{subsec:results_pol_models_comparison}

\begin{figure}[t]
	\centering
	\includegraphics[width=1\linewidth]{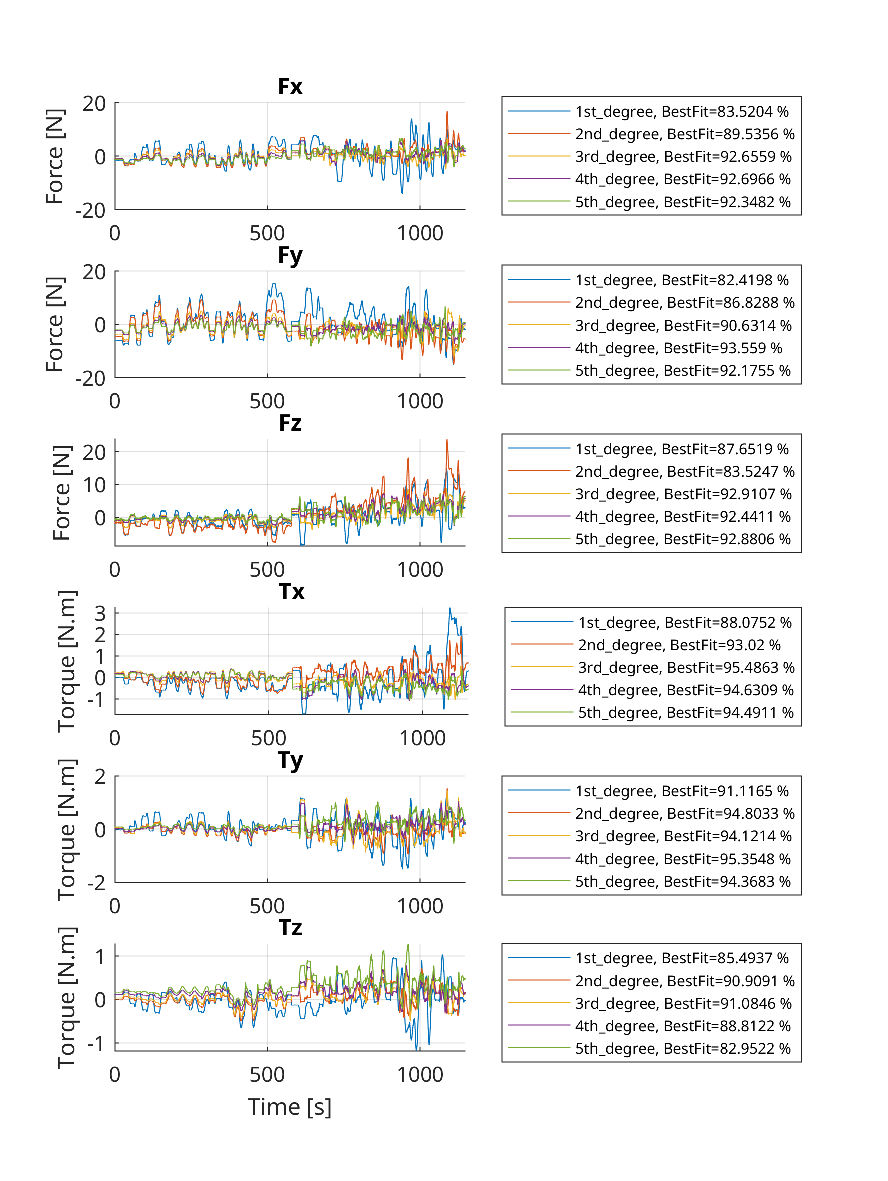}
	\caption{Validation error of the considered models. The \emph{BestFit} criteria is the Normalized Mean Squared Error (NMSE), given by $BestFit = 1-\sqrt{ \frac{MSE}{ \frac{1}{N} \sum_{t=1} (y(t)-\bar{y})^2 } }$, where $MSE = \frac{1}{N} \sum_{t=1}^{N} ( y(t) - \hat{y} (t) )^2$}
	\label{fig:results_validation_error}
\end{figure}

The datasets from the trajectories described in the previous section~\ref{sec:experiments}, are combined, then divided into \emph{training} and \emph{validation} parts. The datasets consist of pairs of expectation (output), and measurement (input), as explained in section~\ref{sec:modeling}. The models tested with the datatsets are detailed in Table~\ref{tab:models_params_rmse}.
Fig.~\ref{fig:results_validation_error} shows the error between the expected force/torque values and the predicted values by each of the tested models.

\begin{figure}[t]
	\centering
	\includegraphics[width=1\linewidth]{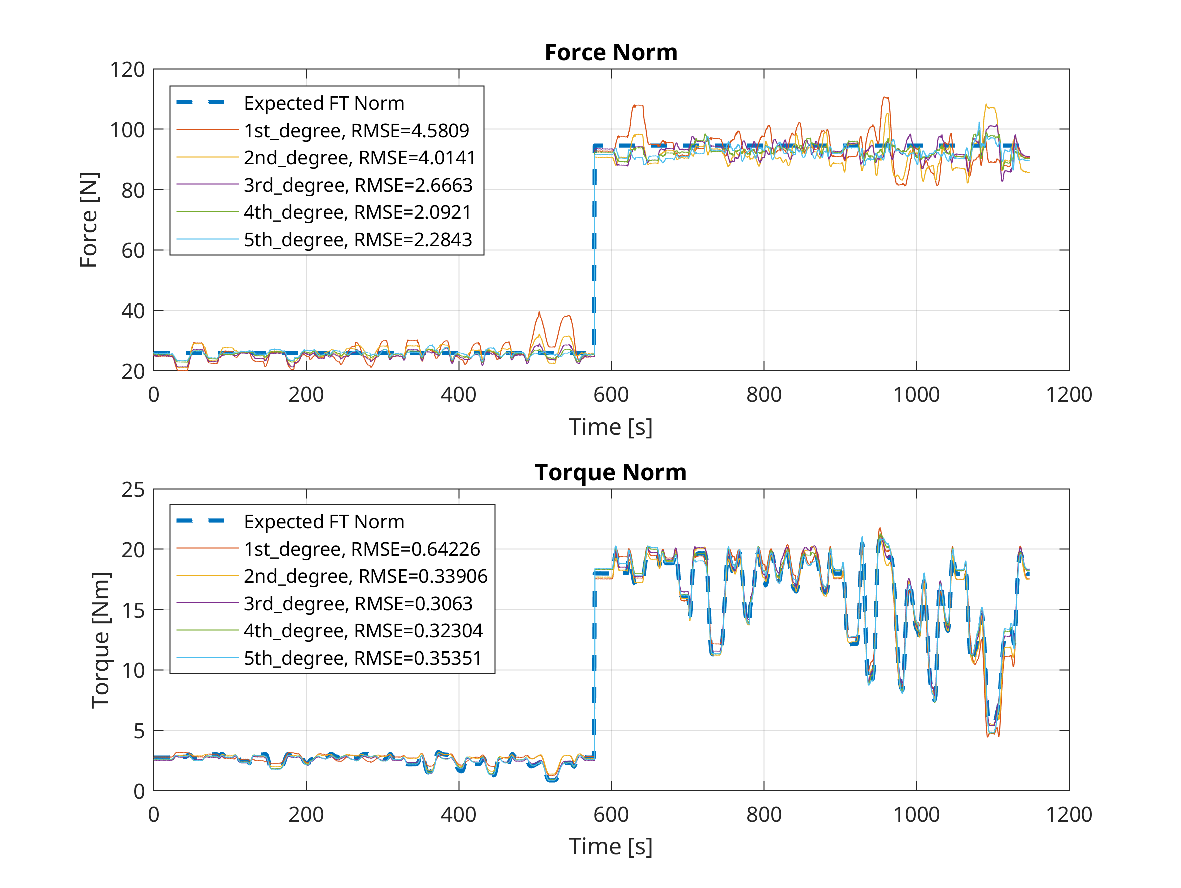}
	\caption{Plots showing the force and torque norms of the considered models against the expected values, considering the validation dataset.}
	\label{fig:results_validation_norm}
\end{figure}

Fig.~\ref{fig:results_validation_error} indicates an improvement of the higher degree models over the affine model. Few Newtons (or Nm) of error are slightly corrected from each force (or torque) component. Plotting the norm of the expected forces and torques against the norm of the estimates of each model (Fig.~\ref{fig:results_validation_norm}), better highlights the improvement of the estimation achieved with higher degree models. Since the norm of the force remains constant for each weight group, the estimates of each model is fluctuating around the expected value. While the fluctuations of the higher degree models are kept around 6N, the affine model's errors on the force norms can reach up to 16N. Further, comparing the RMSE values, we see a significant drop from 4.58N with the affine model, to 2.09N with the 4th-degree model.\looseness=-1

It is also interesting to see in Table~\ref{tab:models_params_rmse} that while the RMSE of the 5th-degree model is slightly less than the RMSE of the 4th-degree model in the training dataset, the validation dataset shows a slight increase in RMSE. This can be an indicator that the 5th-degree model is overfitting the training data. This trend of the RMSE against the model complexity is similar to the one often considered in Machine learning literature, widely known as the \emph{Validation curve}~\cite{goodfellow2016deep}.\looseness=-1

\begin{table}[t]
\caption{Values of the parameters defined in ~\ref{sec:modeling} for the models considered in the experiment, and the corresponding Root Mean Square Error (RMSE) values.}
\vspace{-0.45cm}
\label{tab:models_params_rmse}
\begin{center}
\begin{tabular}{ |c|c|c|c|c|c| } 
\hline
\textbf{No.} & \textbf{1} & \textbf{2} & \textbf{3} & \textbf{4} & \textbf{5} \\
\hline
Polynomial     &        &         &         &         &         \\
degree         & 1st & 2nd & 3rd & 4th & 5th \\
\hline
$np$ & 1 & 2 & 3 & 4 & 5\\
$ny$ & 6 & 6 & 6 & 6 & 6 \\
$nu$ & 7 & 7 & 7 & 7 & 7\\
\hline
No. coeff. & 48 & 216 & 720 & 1980 & 4752 \\
\hline
RMSE of        &        &         &         &         &         \\
force norm     &        &         &         &         &         \\
by training    &        &         &         &         &         \\
dataset [N]    & 5.0423 & 3.1166  & 2.0531  & 1.4565  & 1.1809  \\
\hline
RMSE of        &        &         &         &         &         \\
force norm     &        &         &         &         &         \\
by validation  &        &         &         &         &         \\
dataset [N]    & 4.5809 & 4.0141  & 2.6663  & 2.0921  & 2.2843  \\

\hline
\end{tabular}
\end{center}
\end{table}

\subsection{The effect of $l^1$-norm regularization}
\label{subsec:results_lasso}

\begin{table}[t]
    \caption{Values of the parameters defined in ~\ref{sec:modeling} for the 4th-degree polynomial models identified using~\ref{subsec:algorithm_lasso} in experiment~\ref{sec:experiments}, with different values of $\lambda$, and their corresponding Root Mean Square Error (RMSE) values.}
\vspace{-0.45cm}
\label{tab:models_params_rmse_lasso}
\begin{center}
\begin{tabular}{ |c|c|c|c|c|c|c| } 
\hline
\textbf{No.} & \textbf{1} & \textbf{2} & \textbf{3} & \textbf{4} & \textbf{5} & \textbf{6} \\
\hline
Polynomial     &        &         &         &         &         &         \\
degree         & 4th & 4th & 4th & 4th & 4th & 4th \\
\hline
\textbf{$\lambda$} & \textbf{0.5} & \textbf{1} & \textbf{10} & \textbf{50} & \textbf{100} & \textbf{200}\\
\hline
No. coeff. & 1980 & 1980 & 1980 & 1980 & 1980 & 1980 \\
\hline
No. coeff.      &        &         &         &         &         &         \\
$>10^{-9}$   & 1597 & 1488 & 1152 & 1094 & 1037 & 896 \\
\hline
RMSE           &        &         &         &         &         &         \\
of             &        &         &         &         &         &         \\
force norm     &        &         &         &         &         &         \\
by training    &        &         &         &         &         &         \\
dataset [N]    & 0.8293 & 0.878   & 1.124   & 1.3553  & 1.5064  &  2.01   \\
\hline
RMSE           &        &         &         &         &         &         \\
of             &        &         &         &         &         &         \\
force norm     &        &         &         &         &         &         \\
by             &        &         &         &         &         &         \\
validation     &        &         &         &         &         &         \\
dataset [N]    & 7.0591 & 4.2     & 1.6892  & 1.615   & 1.7031  &  2.471  \\

\hline
\end{tabular}
\end{center}
\end{table}

\begin{figure}[t]
	\centering
	\includegraphics[width=1\linewidth]{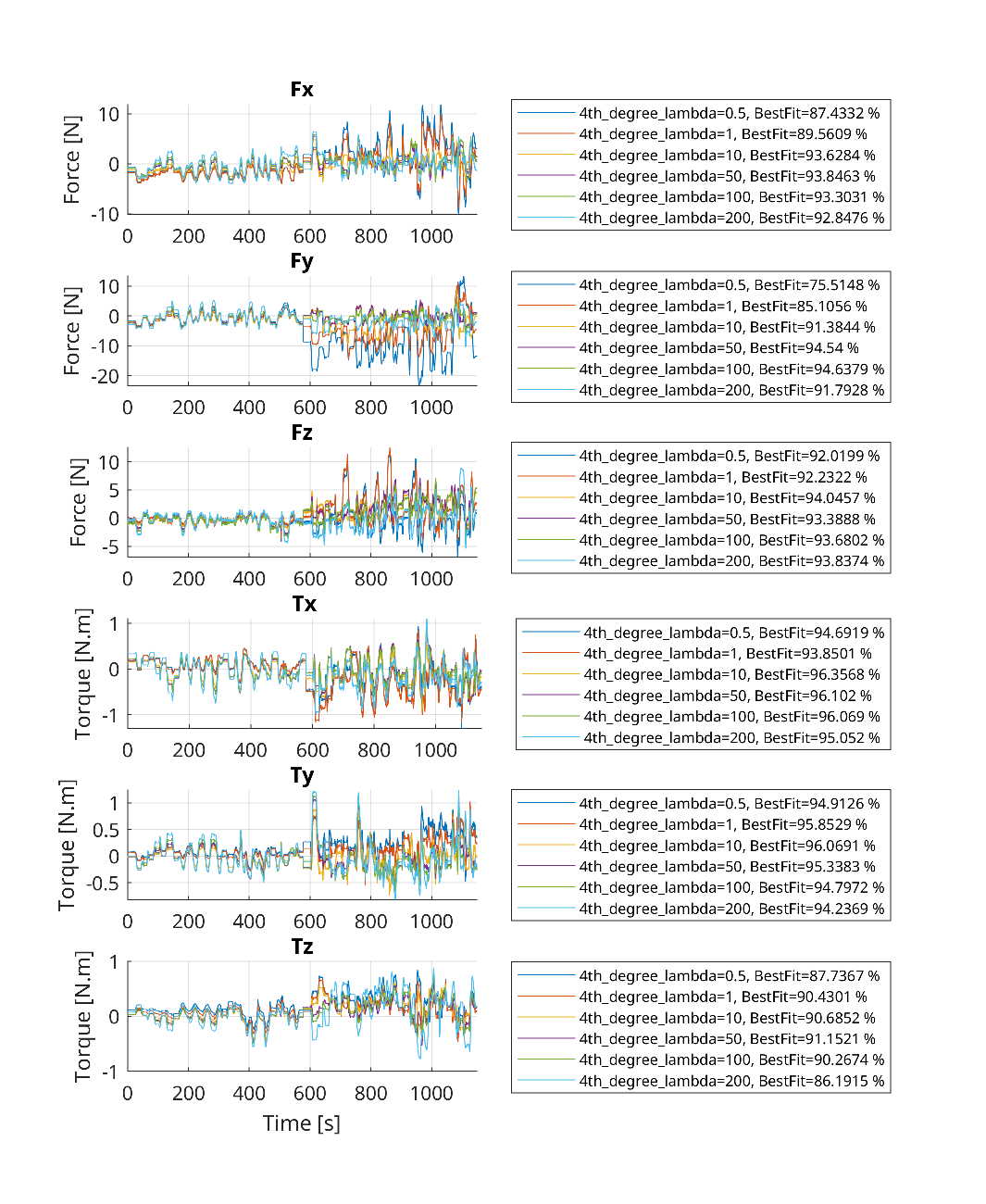}
    \caption{Validation error of the considered 4th-degree polynomial models built using~\ref{subsec:algorithm_lasso}, with different values of the $l^1$-norm regularization weight $\lambda$. Showing the "BestFit" performance measure explained in~\ref{fig:results_validation_error}}
	\label{fig:results_validation_error_lasso}
\end{figure}

\begin{figure}[t]
	\centering
	\includegraphics[width=1\linewidth]{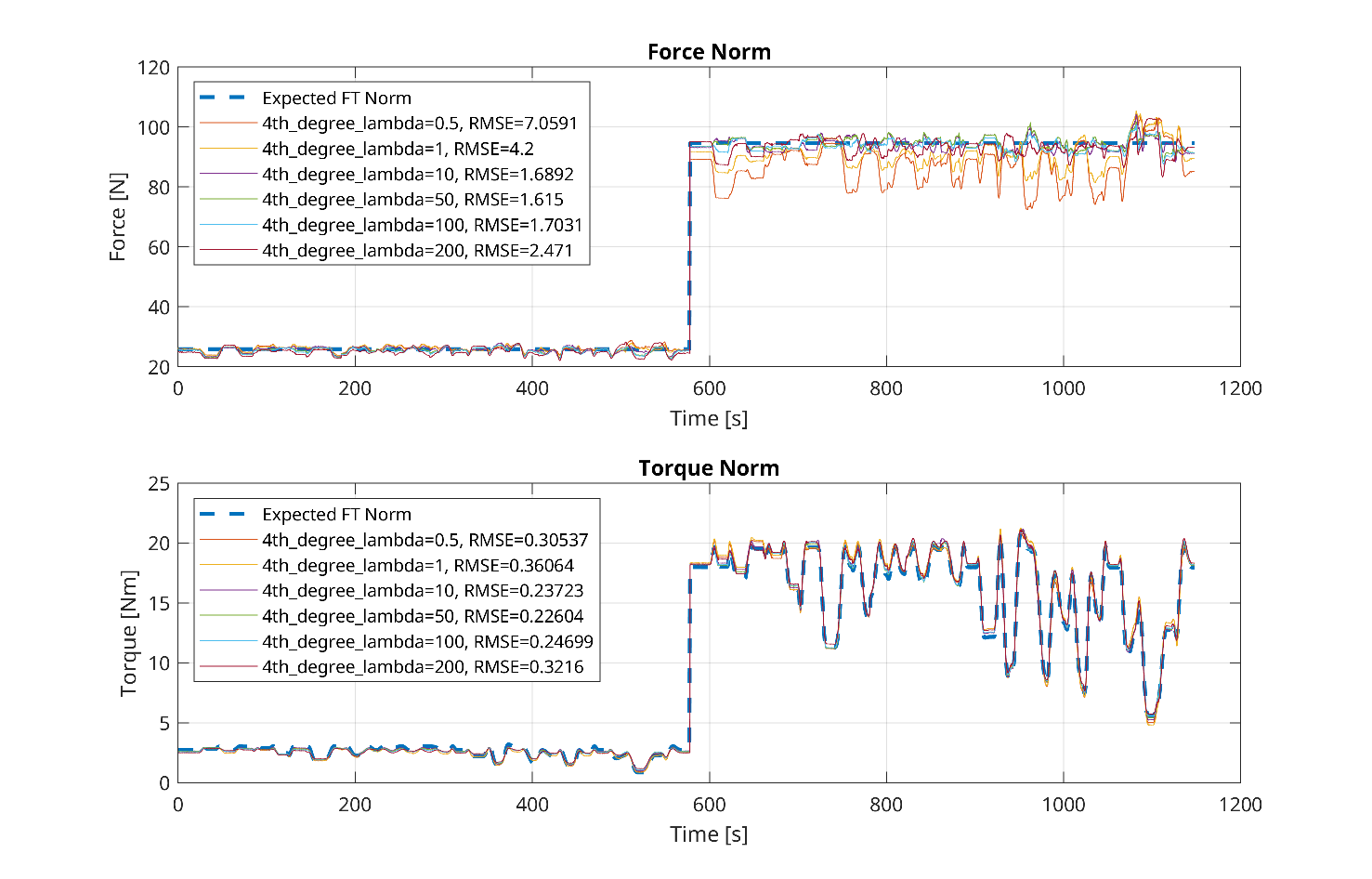}
    \caption{Plots showing the force and torque norms of the considered 4th-degree polynomial models built using~\ref{subsec:algorithm_lasso}, against the expected values, considering the validation dataset, similar to~\ref{sec:experiments}}
	\label{fig:results_validation_norm_lasso}
\end{figure}

With the same dataset used in~\ref{subsec:results_pol_models_comparison}, a group of 4th-degree polynomial models were built, but this time using the algorithm in ~\ref{subsec:algorithm_lasso} to enforce sparsity on the obtained parameters.\looseness=-1

We identified the same model class (4th-degree polynomial), but with choosing a different value of $\lambda$ in each case. Table~\ref{tab:models_params_rmse_lasso} shows the selected values of $\lambda$.\looseness=-1

Fig.~\ref{fig:results_validation_error_lasso} shows the error between the expected force/torque values and the predicted values by each of the tested models.
Fig.~\ref{fig:results_validation_error_lasso} demonstrates the effect of $\lambda$ on the estimation accuracy of the calibrated models. Similar to \ref{subsec:results_pol_models_comparison}, we plot the norm of the expected forces and torques against the norm of the estimates of each model (Fig.~\ref{fig:results_validation_norm_lasso}). In this test, the fluctuations of estimated force norms around the expected values are kept around 5N when $\lambda=50$, which is even better than the 4th-degree model estimated using~\eqref{eq:qp}, with an RMSE of 1.6N compared to 2.09N obtained by latter.

By also looking at the values of the coefficients of the identified 4th-degree polynomial models, we can note from Table~\ref{tab:models_params_rmse_lasso} that the more we increase the value of $\lambda$, the less number of coefficients larger than a threshold $10^{-9}$ becomes. Which gives the number of effective coefficients. The reason this particular value of threshold is selected, is because neglecting any identified coefficient with value less than that threshold did not affect the estimation of the model. This result is a step in the direction of increasing the estimation accuracy while reducing the model complexity.\looseness=-1

\section{CONCLUSIONS}
\label{sec:conclusions}

This work attempts to improve the measurement accuracy of strain-gauge F/T sensors to make them suitable for tasks that span wider force/torque ranges, taking also the temperature variations into account, such tasks like estimating the thrust intensities of flying humanoid robots. When performing the calibration around a smaller range than what was tested in~\ref{sec:experiments}, the affine model~\ref{eq:lin_model} is capable of providing fairly accurate estimates. However, the affine model shows limitations when testing the F/T with wider ranges of forces and torques.
%
It was shown in~\ref{sec:results} that higher-degree polynomial models can improve the estimation accuracy of around $54.33\%$ in RMSE of the norm of the expected forces against the estimates. The significance of this improvement depends on the application for which the sensor is going to be used, and to what extent this error would hinder the performance of the accompanying controller.\looseness=-1

Using Neural Network models can be another viable option, however, polynomial models result in more understandable models. Moreover, Neural Networks often require larger datasets with even more coverage. \looseness=-1

It's worth mentioning that the polynomial model explained in~\eqref{eq:pol_combinations} can be easily extended to represent the system dynamics - if present - in the sense of ARX models~\cite{ljung2017sysid}. This can be done by considering also the previous output and input samples in the model, which then can be written as the following:

\begin{multline}
    \label{eq:pol_arx}
    y_i(k) = \sum_{j=1}^{na} \sum_{l=1}^{ny} \sigma_{ijl} \, y_l(k-j) + \\ \sum_{j=0}^{nb} \sum_{l=1}^{nu} \alpha_{ijl} \, u_l(k-j) + \underbrace{\sum_{j=0}^{nb} \sum_{l,m \in \mathcal{G}} \beta_{ijlm} \, u_l(k-j) u_m(k-j)}_{Quadratic terms} + \\ \underbrace{\sum_{j=0}^{nb} \sum_{l,m,q \in \mathcal{K}} \gamma_{ijlmq} \, u_l(k-j) u_m(k-j) u_q(k-j)}_{Cubic terms} + ... + o_i ,
\end{multline}

Introducing the coefficients $\sigma$ for the previous output $\vect{y}$ samples, and augmenting $\alpha$, $\beta$ and $\gamma$ to include also the past samples of the inputs $\vect{u}$. Also, similar to $np$, $ny$ and $nu$, this formulation introduces the parameters $na$ to represent the order of the dynamics, and $nb$ to represent the system delay.

We also tested the formulation~\eqref{eq:pol_arx} with the dataset collected in~\ref{sec:experiments}, selecting various values of $na$ and $nb$, but it did not show any significant improvement in estimation accuracy. Therefore, we adopted only instantaneous models, as explained in~\ref{sec:results}. This could be an indication that the F/T sensors we are using have dynamics fast enough to be neglected.


Future work includes developing new configurations to allow covering more regions in the space of possible force/torque values applied on each sensor.
Further, validating the proposed calibration models by testing the performance of a closed-loop controller that uses the calibrated F/T sensor's measurements is also a future direction. Which could be the iRonCub flight controller that uses estimated thrust intensities from F/T sensor, or the existing iCub torque control, whose joint torques values are estimated from the F/T sensor's measurements.\looseness=-1
Moreover, it would be interesting to test different models of strain-gauge F/T sensors, in order to study if their design would have an effect on the choice of the calibration model. Further, comparing the performance of higher-degree polynomial models with Neural Networks would more effectively highlight the efficacy of the proposed models.\looseness=-1


\bibliographystyle{IEEEtran}
\bibliography{IEEEabrv,Biblio}

\begin{thebibliography}{10}
\providecommand{\url}[1]{#1}
\csname url@rmstyle\endcsname
\providecommand{\newblock}{\relax}
\providecommand{\bibinfo}[2]{#2}
\providecommand\BIBentrySTDinterwordspacing{\spaceskip=0pt\relax}
\providecommand\BIBentryALTinterwordstretchfactor{4}
\providecommand\BIBentryALTinterwordspacing{\spaceskip=\fontdimen2\font plus
\BIBentryALTinterwordstretchfactor\fontdimen3\font minus \fontdimen4\font\relax}
\providecommand\BIBforeignlanguage[2]{{%
\expandafter\ifx\csname l@#1\endcsname\relax
\typeout{** WARNING: IEEEtran.bst: No hyphenation pattern has been}%
\typeout{** loaded for the language `#1'. Using the pattern for}%
\typeout{** the default language instead.}%
\else
\language=\csname l@#1\endcsname
\fi
#2}}

\bibitem{goswami2019humanoid}
A.~Goswami and P.~Vadakkepat, \emph{Humanoid robotics: a reference}.\hskip 1em plus 0.5em minus 0.4em\relax Springer, 2019.

\bibitem{nava2018position}
G.~{Nava}, L.~{Fiorio}, S.~{Traversaro}, and D.~{Pucci}, ``Position and attitude control of an underactuated flying humanoid robot\looseness=-1,'' in \emph{2018 IEEE-RAS 18th Int. Conf. on Humanoid Robots (Humanoids)\looseness=-1}, 2018, pp. 1--9.

\bibitem{Bartolozzi2017icub}
\BIBentryALTinterwordspacing
L.~Natale, C.~Bartolozzi, D.~Pucci, A.~Wykowska, and G.~Metta, ``icub: The not-yet-finished story of building a robot child,'' \emph{Science Robotics}, vol.~2, no.~13, 2017. [Online]. Available: \url{https://robotics.sciencemag.org/content/2/13/eaaq1026}
\BIBentrySTDinterwordspacing

\bibitem{stasse2017talos}
O.~Stasse, T.~Flayols, R.~Budhiraja, K.~Giraud-Esclasse, J.~Carpentier, J.~Mirabel, A.~Del~Prete, P.~Souères, N.~Mansard, F.~Lamiraux, J.-P. Laumond, L.~Marchionni, H.~Tome, and F.~Ferro, ``Talos: A new humanoid research platform targeted for industrial applications,'' in \emph{2017 IEEE-RAS 17th International Conference on Humanoid Robotics (Humanoids)}, 2017, pp. 689--695.

\bibitem{negrello2016walkman}
F.~Negrello, M.~Garabini, M.~Catalano, P.~Kryczka, W.~Choi, D.~Caldwell, A.~Bicchi, and N.~Tsagarakis, ``Walk-man humanoid lower body design optimization for enhanced physical performance,'' in \emph{2016 IEEE International Conference on Robotics and Automation (ICRA)}, 2016, pp. 1817--1824.

\bibitem{barlian2009piezoresistance}
A.~A. Barlian, W.-T. Park, J.~R. Mallon, A.~J. Rastegar, and B.~L. Pruitt, ``Review: Semiconductor piezoresistance for microsystems,'' \emph{Proceedings of the IEEE}, vol.~97, no.~3, pp. 513--552, 2009.

\bibitem{chavez2019insitu}
\BIBentryALTinterwordspacing
F.~J. Andrade~Chavez, S.~Traversaro, and D.~Pucci, ``Six-axis force torque sensor model-based in situ calibration method and its impact in floating-based robot dynamic performance,'' \emph{Sensors}, vol.~19, no.~24, 2019. [Online]. Available: \url{https://www.mdpi.com/1424-8220/19/24/5521}
\BIBentrySTDinterwordspacing

\bibitem{ati20xxdatasheet}
\emph{Six-Axis Force/Torque Sensor System}, \url{http://www.ati-ia.com/es-MX/app_content/documents/9610-05-1001%20CTL.pdf}, ATI Industrial Automation.

\bibitem{robotiqs2019datasheet}
\BIBentryALTinterwordspacing
\emph{Robotiq FT 300 Force Torque Sensor}, Robotiq, 2019. [Online]. Available: \url{https://assets.robotiq.com/website-assets/support_documents/document/FT_Sensor_Instruction_Manual_PDF_20190322.pdf}
\BIBentrySTDinterwordspacing

\bibitem{chavez2019temperature}
F.~J.~A. Chavez, G.~Nava, S.~Traversaro, F.~Nori, and D.~Pucci, ``Model based in situ calibration with temperature compensation of 6 axis force torque sensors,'' in \emph{2019 International Conference on Robotics and Automation (ICRA)}, 2019, pp. 5397--5403.

\bibitem{Mohamed2022thrust}
H.~A.~O. Mohamed, G.~Nava, G.~L’Erario, S.~Traversaro, F.~Bergonti, L.~Fiorio, P.~R. Vanteddu, F.~Braghin, and D.~Pucci, ``Momentum-based extended kalman filter for thrust estimation on flying multibody robots,'' \emph{IEEE Robotics and Automation Letters}, vol.~7, no.~1, pp. 526--533, 2022.

\bibitem{traversaro2015insitu}
S.~Traversaro, D.~Pucci, and F.~Nori, ``In situ calibration of six-axis force-torque sensors using accelerometer measurements,'' in \emph{2015 IEEE International Conference on Robotics and Automation (ICRA)}, 2015, pp. 2111--2116.

\bibitem{ding2022insitu}
C.~Ding, Y.~Han, W.~Du, J.~Wu, and Z.~Xiong, ``In situ calibration of six-axis force–torque sensors for industrial robots with tilting base,'' \emph{IEEE Transactions on Robotics}, vol.~38, no.~4, pp. 2308--2321, 2022.

\bibitem{oh2018deeplearning}
H.~S. Oh, U.~Kim, G.~Kang, J.~K. Seo, and H.~R. Choi, ``Multi-axial force/torque sensor calibration method based on deep-learning,'' \emph{IEEE Sensors Journal}, vol.~18, no.~13, pp. 5485--5496, 2018.

\bibitem{su2019modelfree}
\BIBentryALTinterwordspacing
H.~Su, W.~Qi, Y.~Hu, J.~Sandoval, L.~Zhang, Y.~Schmirander, G.~Chen, A.~Aliverti, A.~Knoll, G.~Ferrigno, and E.~De~Momi, ``Towards model-free tool dynamic identification and calibration using multi-layer neural network,'' \emph{Sensors}, vol.~19, no.~17, 2019. [Online]. Available: \url{https://www.mdpi.com/1424-8220/19/17/3636}
\BIBentrySTDinterwordspacing

\bibitem{tienfu1997neural}
T.-F. Lu, G.~C. Lin, and J.~R. He, ``Neural-network-based 3d force/torque sensor calibration for robot applications,'' \emph{Engineering Applications of Artificial Intelligence}, vol.~10, no.~1, pp. 87--97, 1997.

\bibitem{stellato2020osqp}
\BIBentryALTinterwordspacing
B.~Stellato, G.~Banjac, P.~Goulart, A.~Bemporad, and S.~Boyd, ``{OSQP}: an operator splitting solver for quadratic programs,'' \emph{Mathematical Programming Computation}, vol.~12, no.~4, pp. 637--672, 2020. [Online]. Available: \url{https://doi.org/10.1007/s12532-020-00179-2}
\BIBentrySTDinterwordspacing

\bibitem{tibshirani1996lasso}
R.~Tibshirani, ``Regression shrinkage and selection via the lasso,'' \emph{Journal of the Royal Statistical Society Series B: Statistical Methodology}, vol.~58, no.~1, pp. 267--288, 1996.

\bibitem{dafarra2022icub3}
S.~Dafarra, K.~Darvish, R.~Grieco, G.~Milani, U.~Pattacini, L.~Rapetti, G.~Romualdi, M.~Salvi, A.~Scalzo, I.~Sorrentino, D.~Tomè, S.~Traversaro, E.~Valli, P.~M. Viceconte, G.~Metta, M.~Maggiali, and D.~Pucci, ``icub3 avatar system,'' 2022.

\bibitem{pucci2017fly}
D.~Pucci, S.~Traversaro, and F.~Nori, ``Momentum control of an underactuated flying humanoid robot,'' \emph{IEEE Robotics and Automation Letters}, vol.~3, no.~1, pp. 195--202, Jan 2018.

\bibitem{icub-tech}
\BIBentryALTinterwordspacing
``icub tech facility in the italian institute of technology - genoa, italy.'' [Online]. Available: \url{https://github.com/icub-tech-iit}
\BIBentrySTDinterwordspacing

\bibitem{urdf}
\BIBentryALTinterwordspacing
``Xml robot description format (urdf).'' [Online]. Available: \url{http://wiki.ros.org/urdf/XML/model}
\BIBentrySTDinterwordspacing

\bibitem{guedelha2016self}
N.~Guedelha, N.~Kuppuswamy, S.~Traversaro, and F.~Nori, ``Self-calibration of joint offsets for humanoid robots using accelerometer measurements,'' in \emph{Humanoid Robots (Humanoids), 2016 IEEE-RAS 16th International Conference on}.\hskip 1em plus 0.5em minus 0.4em\relax IEEE, 2016, pp. 1233--1238.

\bibitem{nori2015idyntree}
F.~Nori, S.~Traversaro, J.~Eljaik, F.~Romano, A.~Del~Prete, and D.-. Pucci, ``icub whole-body control through force regulation on rigid noncoplanar contacts\looseness=-1,'' \emph{Frontiers in Robotics and AI\looseness=-1}, vol.~2, no.~6, 2015.

\bibitem{traversaro2017modelling}
\BIBentryALTinterwordspacing
S.~{Traversaro}, ``Modelling, estimation and identification of humanoid robots dynamics,'' Ph.D. dissertation, Italian Institute of Technology, 2017. [Online]. Available: \url{https://github.com/traversaro/traversaro-phd-thesis}
\BIBentrySTDinterwordspacing

\bibitem{goodfellow2016deep}
I.~Goodfellow, Y.~Bengio, and A.~Courville, \emph{Deep Learning}.\hskip 1em plus 0.5em minus 0.4em\relax MIT Press, 2016, \url{http://www.deeplearningbook.org}.

\bibitem{ljung2017sysid}
\BIBentryALTinterwordspacing
L.~Ljung, \emph{System Identification}.\hskip 1em plus 0.5em minus 0.4em\relax John Wiley and Sons, Ltd, 2017, pp. 1--19. [Online]. Available: \url{https://onlinelibrary.wiley.com/doi/abs/10.1002/047134608X.W1046.pub2}
\BIBentrySTDinterwordspacing

\end{thebibliography}

\end{document}